\documentclass[11pt]{article}

\usepackage[margin=1in]{geometry}
\usepackage[utf8]{inputenc}
\usepackage[T1]{fontenc}

\usepackage{amsmath,amssymb}

\usepackage{booktabs}
\usepackage{graphicx}
\usepackage{float}
\usepackage{multirow}
\usepackage{threeparttable}
\usepackage{tabularx}
\usepackage{caption}

\usepackage{listings}
\usepackage{xcolor}
\usepackage[colorlinks=true, linkcolor=blue!70!black, citecolor=blue!70!black, urlcolor=blue!70!black]{hyperref}

\usepackage[numbers,sort&compress]{natbib}

\usepackage{tikz}
\usetikzlibrary{shapes.geometric, arrows.meta, positioning, fit, backgrounds, calc}

\renewcommand{\thefootnote}{\fnsymbol{footnote}}

\title{EDM-ARS: A Domain-Specific Multi-Agent System for \\Automated Educational Data Mining Research}

\author{
  Chenguang Pan \quad
  Zhou Zhang \quad
  Weixuan Xiao \quad
  Chengyuan Yao \\[4pt]
  \textit{edmars.ai} \\
  \textit{New York, NY, USA} \\
  \texttt{\{cp3280, zz2863, wx2299,cy2706\}@tc.columbia.edu}
}

\date{March 2026}

\begin{document}
\maketitle

\renewcommand{\thefootnote}{\arabic{footnote}}

\begin{abstract}
In this technical report, we present the Educational Data Mining Automated Research System (EDM-ARS), a domain-specific multi-agent pipeline that automates end-to-end educational data mining (EDM) research. We conceptualize EDM-ARS as a general framework for domain-aware automated research pipelines, where educational expertise is embedded into each stage of the research lifecycle. As a first instantiation of this framework, we focus on predictive modeling tasks. Within this scope, EDM-ARS orchestrates five specialized LLM-powered agents (ProblemFormulator, DataEngineer, Analyst, Critic, and Writer) through a state-machine coordinator that supports revision loops, checkpoint-based recovery, and sandboxed code execution. Given a research prompt and a dataset, EDM-ARS produces a complete LaTeX manuscript with real Semantic Scholar citations, validated machine learning analyses, and automated methodological peer review. We also provide a detailed description of the system architecture, the three-tier data registry design that encodes educational domain expertise, the specification of each agent, the inter-agent communication protocol, and mechanisms for error-handling and self-correction. Finally, we discuss current limitations, including single-dataset scope and formulaic paper output, and outline a phased roadmap toward causal inference, transfer learning, psychometric, and multi-dataset generalization. EDM-ARS is released as an open-source project to support the educational research community.\footnote{This document is a technical report for EDM-ARS rather than a research paper. For the latest updates, please visit our website: \url{https://edmars.ai/}. The source code is available at \url{https://github.com/cgpan/edm-ars-public}.}
\end{abstract}

\vspace{0.5em}

\section{Introduction}
\label{sec:intro}

Automated scientific research powered by large language models (LLMs) has become an active area of study. Systems such as The AI Scientist~\citep{lu2024aiscientist}, its successor AI Scientist v2~\citep{yamada2025aiscientistv2}, CycleResearcher~\citep{weng2025cycleresearcher}, DeepScientist~\citep{weng2025deepscientist}, AI-Researcher~\citep{tang2025airesearcher}, and Aletheia~\citep{feng2026aletheia} have shown that LLM-based agents can perform literature review, formulate hypotheses, write and execute code, analyze results, and draft manuscripts with minimal human intervention. These systems represent an emerging paradigm in which the scientific discovery loop, from idea generation to paper writing, is largely automated.

Several concerns arise when examining this landscape. A general, domain-agnostic system may aim for generality across scientific fields but consequently lacks the domain knowledge that experienced human researchers bring to specialized areas. For example, a general system that does not know that an educational data survey uses negative sentinel codes for missing data, or that temporal leakage between survey waves invalidates predictive models, will silently produce methodologically flawed research. The computational cost of these systems is also high; general-purpose agentic search strategies (e.g., the tree-search approach of AI Scientist v2) require many LLM calls, and the token consumption per paper can be impractical for researchers without institutional API budgets. The quality of generated papers remains debated, with many outputs exhibiting formulaic structure, shallow interpretation, and hallucinated citations. And despite the large and growing body of educational data mining (EDM) and learning analytics (LA) research~\citep{romero2020edm}, virtually no automated research system has targeted the educational research community.

To address these gaps, we present EDM-ARS (Educational Data Mining Automated Research System), a domain-specific multi-agent pipeline that automates end-to-end EDM research, beginning with prediction tasks as its first fully supported paradigm. EDM-ARS is built on three premises. A portion of EDM research follows standardized, well-defined methodological workflows: problem formulation from a dataset codebook, data cleaning with survey-specific conventions, model training with interpretability analysis, fairness-aware evaluation, and structured academic writing. These workflows are sufficiently regular to be captured by a pipeline of specialized agents. Domain knowledge is treated as a first-class design element rather than an afterthought: the system's data registry, validation rules, and agent prompts encode accumulated expertise in working with large-scale educational survey data. And by constraining the system's scope to a well-defined research paradigm (prediction on a single representative dataset for the current version), we aim to achieve higher reliability and output quality than a general-purpose system attempting to cover all of science.

As a first instantiation of this pipeline, EDM-ARS currently focuses on prediction tasks using the High School Longitudinal Study of 2009 (HSLS:09), a nationally representative longitudinal survey conducted by the National Center for Education Statistics (NCES) tracking approximately 23,500 students from 9th grade (2009) through postsecondary education (2016). 

Beyond this first instantiation, we plan to expand EDM-ARS to support other educational data mining tasks in future versions includes causal inference (quasi-experimental designs, heterogeneous treatment-effect analysis, optimal treatment regimes), transfer learning (cross-domain and cross-population generalization), and psychometric applications.

This project began as a hobby project, but we treat it with the rigor of a production system. The remainder of this report is organized as follows. Section~\ref{sec:related} surveys the current landscape of automated research systems and positions EDM-ARS within it. Section~\ref{sec:architecture} describes the pipeline architecture, the orchestrator state machine, design rationale, and technology stack. Section~\ref{sec:registry} details the data registry, including the three-tier variable strategy, domain knowledge encoding, and leakage prevention mechanisms. Section~\ref{sec:agents} specifies each of the five agents. Section~\ref{sec:communication} describes inter-agent communication protocols. Section~\ref{sec:evaluation} outlines the evaluation framework. Section~\ref{sec:limitations} discusses current limitations and future work. Section~\ref{sec:reproducibility} provides deployment and usage instructions.

\section{Related Work: Automated Research Systems}
\label{sec:related}

The landscape of LLM-powered automated research has evolved rapidly since 2024. We organize the discussion around general-purpose automated research systems, domain-specific automation, and educational data mining.

\subsection{General-Purpose Automated Research Systems}

The AI Scientist~\citep{lu2024aiscientist} was among the first end-to-end systems to automate the full research lifecycle: idea generation, experiment execution, manuscript writing, and simulated peer review. Operating within predefined code templates (e.g., NanoGPT, 2D diffusion), the system used an LLM to iteratively modify experimental code, collect results, and write a paper. Its main limitation was the reliance on human-authored code scaffolds, which constrained the system to domains where such templates existed.

The AI Scientist v2~\citep{yamada2025aiscientistv2} addressed this constraint through an agentic tree search architecture. Rather than following a single linear pipeline, v2 explores multiple experimental branches in parallel, managed by a dedicated Experiment Manager agent. Different branches produce different manuscripts, and the best is selected via automated review. The system also eliminated reliance on human code templates by generating experimental code from scratch, and introduced a Vision-Language Model (VLM) feedback loop for iterative figure quality improvement. AI Scientist v2 demonstrated workshop-level paper quality, though at high computational cost due to the breadth-first exploration strategy.

CycleResearcher~\citep{weng2025cycleresearcher} introduced an iterative reinforcement learning framework for automated research. CycleResearcher (the policy model) generates papers, while CycleReviewer (the reward model) scores them. Preference pairs are constructed from multiple samples, and the policy is refined via SimPO (Simple Preference Optimization). This feedback loop allows the system to learn from its own review signals, producing progressively better papers. CycleResearcher differs from single-shot pipelines in a fundamental way: rather than relying on a fixed prompt to produce quality, it uses learned optimization to improve output over time.

DeepScientist~\citep{weng2025deepscientist} extended this direction to frontier-level scientific findings. The system maintains a cumulative knowledge base across runs, so each new research question is informed by prior outputs. This addresses a limitation of amnesic systems that treat each run independently and may redundantly explore the same research space. DeepScientist's progressive knowledge accumulation allows it to build on its own prior findings, much as a research group develops expertise over a series of related publications.

AI-Researcher~\citep{tang2025airesearcher} emphasized autonomous scientific innovation, incorporating automated idea validation, experiment design, and result analysis within a multi-agent framework. The system uses specialized agents for different phases of the research process and includes mechanisms for novelty assessment and contribution evaluation.

Aletheia~\citep{feng2026aletheia} targets autonomous mathematics research, using web search and browsing to navigate mathematical literature. Aletheia showed that domain-specific tool use (e.g., automated theorem provers, symbolic computation) can improve the quality of generated research in formal domains.

Two additional efforts are worth noting. FARS (Fully Autonomous Research
System)~\citep{fars2026} is a commercial automated research platform targeting
general scientific domains. AutoResearchClaw~\citep{autoresearchclaw2026} is an
open-source framework for automated research that emphasizes citation
verification and literature grounding.

\subsection{Domain-Specific Considerations}

A recurring tension across general-purpose systems is the tradeoff between generality and depth. Systems designed to work across all scientific domains necessarily lack the specialized knowledge that makes research in any particular field methodologically sound. In educational data mining, domain knowledge is not a luxury but a prerequisite for producing valid research.

The reasons are concrete. NCES survey datasets encode missing data using negative sentinel codes ($-1, -4, -7, -8, -9$) with distinct substantive meanings (legitimate skip, nonrespondent, missing by design, etc.); a system that treats these as valid numeric values will produce silently corrupted analyses. Variables in longitudinal studies have temporal ordering constraints, and using 11th-grade outcomes to predict 9th-grade characteristics constitutes data leakage that inflates model performance. Survey weights exist to support population-level inference; ignoring them in machine learning pipelines is acceptable for prediction but must be explicitly disclosed as a limitation. Protected attributes (race, gender, socioeconomic status) require fairness diagnostics when used in predictive models, and subgroup sample sizes must be assessed before drawing fairness conclusions.

No existing automated research system encodes this kind of domain knowledge. EDM-ARS was designed from the ground up to make such knowledge a structural component of the system rather than a post-hoc addition.

\subsection{Educational Data Mining Research}

Educational data mining and learning analytics constitute a mature field with established methodological standards~\citep{romero2020edm}. Prediction is one of the most common research paradigms in EDM, typically involving the identification of student-level predictors of academic outcomes (achievement, persistence, graduation, enrollment) using supervised machine learning. The field has established norms around model interpretability, including feature importance and SHAP (SHapley Additive exPlanations) analysis, fairness evaluation (subgroup performance disparities), and transparent reporting of data preprocessing decisions.

Despite this methodological regularity, which makes EDM research well-suited to automation, no prior system has been built for the educational research community. EDM-ARS is a first step in this direction, encoding EDM-specific methodological knowledge and currently operating on the HSLS:09 dataset, with plans to expand to additional educational datasets such as IPEDS, PISA, and ASSISTments.

\section{EDM-ARS Architecture}
\label{sec:architecture}

\subsection{High-Level Pipeline}
\label{sec:pipeline}

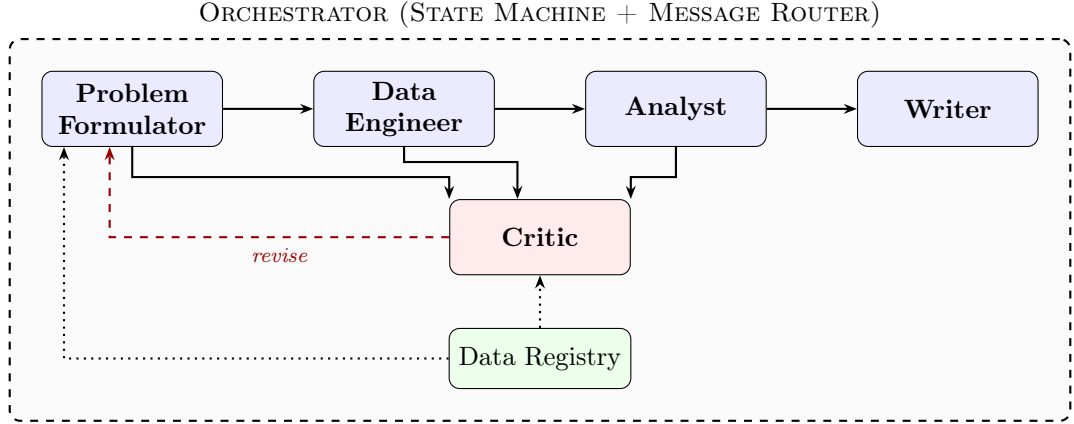
\begin{figure}[!h]
\centering
\begin{tikzpicture}[
    node distance=0.5cm and 1.2cm,
    agent/.style={draw, rounded corners, fill=blue!8, minimum width=2.4cm, minimum height=1cm, align=center, font=\small\bfseries},
    data/.style={draw, rounded corners, fill=green!8, minimum width=2.2cm, minimum height=0.8cm, align=center, font=\small},
    orch/.style={draw, thick, dashed, rounded corners, inner sep=12pt},
    arr/.style={-{Stealth[length=5pt]}, thick},
]
  \node[agent] (pf) {Problem\\Formulator};
  \node[agent, right=of pf] (de) {Data\\Engineer};
  \node[agent, right=of de] (an) {Analyst};
  \node[agent, right=of an] (wr) {Writer};

  \node[agent, fill=red!8, below=1.2cm of $(de)!0.5!(an)$] (cr) {Critic};

  \node[data, below=0.7cm of cr] (dr) {Data Registry};

  \draw[arr] (pf) -- (de);
  \draw[arr] (de) -- (an);
  \draw[arr] (an) -- (wr);

  \draw[arr] (pf.south) -- ++(0,-0.4) -| (cr.north west);
  \draw[arr] (de.south) -- ++(0,-0.2) -| ([xshift=-0.3cm]cr.north);
  \draw[arr] (an.south) -- ++(0,-0.4) -| (cr.north east);

  \draw[arr, dashed, red!60!black] (cr.west) -| ([xshift=-0.3cm]pf.south) node[pos=0.25, below, font=\scriptsize\itshape] {revise};

  \draw[arr, dotted] (dr) -- (cr);
  \draw[arr, dotted] (dr.west) -| ([xshift=0.3cm]pf.south west);

  \begin{scope}[on background layer]
    \node[orch, fill=gray!4, fit=(pf)(de)(an)(wr)(cr)(dr), label={[font=\small\scshape]above:Orchestrator (State Machine + Message Router)}] {};
  \end{scope}
\end{tikzpicture}
\caption{EDM-ARS architecture. Five specialized agents are coordinated by a state-machine orchestrator. The Critic agent reviews outputs from the ProblemFormulator, DataEngineer, and Analyst, and can issue revision instructions routed back through the pipeline. The Data Registry provides domain knowledge to all agents. Solid arrows indicate the forward pipeline; dashed arrows indicate revision loops; dotted arrows indicate domain knowledge access.}
\label{fig:pipeline}
\end{figure}

EDM-ARS implements a five-agent sequential pipeline coordinated by a central orchestrator. The pipeline takes as input a research prompt (optional) and the HSLS:09 public-use dataset, and produces as output a complete LaTeX manuscript with supporting tables, figures, and bibliography. Figure~\ref{fig:pipeline} illustrates the architecture.

The five agents execute in sequence. The ProblemFormulator searches literature via Semantic Scholar, scopes a research question, selects outcome and predictor variables from the data registry, and produces a structured research specification. The DataEngineer generates and executes Python code to clean the raw HSLS:09 data, handle missing values following the NCES protocol, encode categorical variables, and produce train/test splits. The Analyst generates and executes Python code to train a battery of machine learning models (Logistic Regression, Random Forest, XGBoost, ElasticNet, MLP, StackingEnsemble), compute SHAP feature importance, perform subgroup fairness analysis, and produce result tables and figures. The Critic reviews all prior agent outputs against a multi-dimensional quality rubric covering methodological rigor, data preparation quality, analysis validity, and educational meaningfulness, issuing one of three verdicts: \textsc{pass}, \textsc{revise}, or \textsc{abort}. The Writer compiles all pipeline outputs into a complete LaTeX manuscript using a pre-defined ACM \texttt{sigconf} template with placeholder markers, generating prose for each section and a BibTeX bibliography from Semantic Scholar metadata.

\subsection{Orchestrator State Machine}
\label{sec:orchestrator}

The orchestrator manages pipeline execution through an explicit finite state machine with the following states: \textsc{initialized}, \textsc{formulating}, \textsc{engineering}, \textsc{analyzing}, \textsc{critiquing}, \textsc{revising}, \textsc{writing}, \textsc{completed}, and \textsc{aborted}.

Forward transitions follow the agent sequence: \textsc{formulating} $\to$ \textsc{engineering} $\to$ \textsc{analyzing} $\to$ \textsc{critiquing}. From \textsc{critiquing}, the next transition depends on the Critic's verdict. A \textsc{pass} verdict advances to \textsc{writing} and then \textsc{completed}. A \textsc{revise} verdict transitions to \textsc{revising}, then loops back to \textsc{critiquing}. An \textsc{abort} verdict transitions to \textsc{aborted}. The maximum number of revision cycles is configurable (default: 2). If the maximum is exhausted without a \textsc{pass} verdict, the pipeline proceeds to \textsc{writing} with an \textsc{unverified} flag. Any unrecoverable error, such as data validation failure, JSON parse failure, or an analytic sample below 1{,}000, also triggers an \textsc{aborted} transition.

\paragraph{Topological revision cascade.} When the Critic issues a \textsc{revise} verdict with targeted instructions, the orchestrator determines the lowest-level targeted agent in the dependency graph (\allowbreak\texttt{Problem\hspace{0pt}Formulator} \(\to\) \texttt{Data\hspace{0pt}Engineer} \(\to\) \texttt{Analyst}) and re-runs that agent plus all downstream agents. For example, if the Critic targets only the DataEngineer, both the DataEngineer and Analyst are re-run; if the Critic targets the ProblemFormulator, all three agents are re-run. This ensures that revisions propagate correctly through the dependency chain.

\paragraph{Checkpoint and recovery.} At each state transition, the orchestrator serializes the full pipeline context (including all agent outputs, logs, and metadata) to a \texttt{checkpoint.json} file in the output directory. On startup, if a checkpoint exists, the orchestrator resumes from the recorded state, skipping completed stages. This enables recovery from crashes, API timeouts, and interrupted runs. Users can invoke recovery via the \texttt{--resume} CLI flag.

\subsection{Design Rationale}
\label{sec:rationale}

Several design principles distinguish EDM-ARS from general-purpose automated research systems.

\paragraph{Domain knowledge as structure, not prompt.} General-purpose systems encode domain knowledge (if at all) through prompt instructions that can be ignored or misinterpreted. EDM-ARS instead encodes it structurally through the data registry (Section~\ref{sec:registry}), validation rules enforced programmatically at stage boundaries, and agent-specific system prompts loaded from external YAML files. The distinction is analogous to the difference between asking a researcher ``please remember to check for temporal leakage'' and building a compiler that rejects temporally invalid variable combinations.

\paragraph{Separation of concerns for LLM tasks.} Our development experience showed that LLMs handle prose generation well but struggle with structural boilerplate such as LaTeX preambles, BibTeX formatting, and complex code scaffolds. EDM-ARS separates these concerns accordingly: the Writer agent fills prose into a pre-defined LaTeX template with \texttt{\%\%PLACEHOLDER:SLOT\_NAME\%\%} markers rather than generating LaTeX from scratch; code execution is handled by a sandboxed environment rather than inline LLM output; and configuration (model IDs, paths, hyperparameters) lives in a central \texttt{config.yaml} file rather than being hardcoded in prompts.

\paragraph{Open-loop pipelines fail silently.} The original EDM-ARS prototype used a linear pipeline without validation checkpoints: the ProblemFormulator generated a spec, the DataEngineer cleaned data, the Analyst ran models, and the Writer produced a paper, all without intermediate quality checks. Failures were detected only at the Critic stage (if at all), by which point computation had been wasted. The current design validates outputs at every stage boundary, uses phased execution within agents (so that a SHAP failure does not destroy already-completed model training), and uses an error taxonomy that classifies failures for targeted repair.

\paragraph{No frameworks.} EDM-ARS uses a custom orchestrator rather than third-party agent frameworks (LangChain, LangGraph, CrewAI). Educational research pipelines have well-defined, stable topologies that may not benefit from the dynamic routing capabilities of general-purpose frameworks. A custom orchestrator also gives us precise control over state management, checkpoint serialization, and revision routing that would require extensive customization of any existing framework.

\subsection{Technology Stack}
\label{sec:techstack}

Table~\ref{tab:techstack} summarizes the technology stack.

\begin{table}[H]
\centering
\caption{EDM-ARS technology stack.}
\label{tab:techstack}
\begin{tabular}{ll}
\toprule
Component & Technology \\
\midrule
Language & Python 3.11+ \\
LLM API & Anthropic SDK (Claude Sonnet for most agents, Opus for Critic) \\
ML / Statistics & pandas, scikit-learn, XGBoost, SHAP, matplotlib, seaborn \\
Configuration & PyYAML, central \texttt{config.yaml} \\
Literature retrieval & Semantic Scholar API\\
Code execution sandbox & Docker container (optional; subprocess fallback) \\
Docker SDK & \texttt{docker} Python package $\geq$ 7.0 (host-side, optional) \\
Output format & LaTeX (ACM \texttt{acmart} template, \texttt{sigconf} style) \\
Agent frameworks & None (custom orchestrator) \\
\bottomrule
\end{tabular}
\end{table}

All LLM calls are routed through a \texttt{BaseAgent.call\_llm()} method that reads model identifiers from \texttt{config.yaml}, keeping model selection centralized and easy to modify. The Critic agent uses the most capable available model (Claude Opus) for review quality, while all other agents use Claude Sonnet for cost efficiency. Agent system prompts are stored in external YAML files (\texttt{agent\_prompts/*.yaml}) and are never hardcoded in Python source code.

\section{Data Registry Design}
\label{sec:registry}

The data registry is one of the core design contributions of EDM-ARS. It encodes domain expertise about the HSLS:09 dataset so that LLM agents can reason about variables substantively, understanding their educational meaning, temporal context, missingness patterns, and analytic constraints, rather than treating them as opaque column names.

\subsection{Three-Tier Variable Strategy}
\label{sec:tiers}

EDM-ARS organizes HSLS:09 variables into three tiers that reflect the level of human curation.

\paragraph{Tier 1: Hand-curated variables ($\sim$95 variables).} These are the most commonly used and educationally meaningful variables in HSLS:09, each annotated with a human-written label, variable type (binary, continuous, categorical), value range, data collection wave, approximate missingness percentage, codebook-level category mappings, and substantive notes. Tier 1 entries include both outcome variables (e.g., \texttt{X2TXMTSCOR}: math theta score; \texttt{X4EVRATNDCLG}: ever attended college) and predictor variables organized by substantive category (achievement, attitudes, socioeconomic status, demographic, school context). The ProblemFormulator draws primarily from Tier 1 when designing research specifications.

\paragraph{Tier 2: Auto-generated variables.} These are generated programmatically by a \texttt{generate\_tier2.py} script that reads the raw HSLS:09 data file, excludes Tier 1 and Tier 3 variables, and infers basic metadata (wave from variable prefix, type from value distribution, missingness percentage). Tier 2 variables have metadata but lack hand-curated substantive annotations. The ProblemFormulator may use Tier 2 variables but must provide stronger educational rationale for their inclusion. Tier 2 entries are organized by data collection wave and include inferred data source (student questionnaire, parent questionnaire, teacher questionnaire, composite/derived).

\paragraph{Tier 3: Excluded variables.} These are variables that must never be used as predictors or outcomes. Exclusion is enforced by three kinds of pattern rules: exact matches for survey weights (\texttt{W1STUDENT}, \texttt{W2W1STU}, etc.), student identifiers (\texttt{STU\_ID}), and strata/PSU variables; prefix patterns for BRR replicate weights and flag variables; and suffix patterns for weight-related variable names.

This three-tier strategy balances coverage (Tier 2 provides access to the full variable space) with quality control (Tier 1 prioritizes well-understood variables) and safety (Tier 3 prevents methodological errors).

\subsection{Domain Knowledge Encoding}
\label{sec:domain_knowledge}

The data registry encodes several forms of domain knowledge that would otherwise require human expertise.

\paragraph{Temporal ordering.} HSLS:09 is a longitudinal study with four waves: base year (2009, prefix \texttt{X1}), first follow-up (2012, prefix \texttt{X2}), second follow-up (2013, prefix \texttt{X3}), and update panel (2016, prefix \texttt{X4}). The registry defines a strict temporal ordering (\texttt{base\_year} $<$ \texttt{first\_follow\_up} $<$ \texttt{second\_follow\_up} $<$ \texttt{update\_panel}) and annotates each variable with its wave. This ordering is used to enforce the constraint that all predictors must temporally precede the outcome variable.

\paragraph{Missing data conventions.} NCES surveys use negative sentinel codes with distinct substantive meanings: $-1$ (legitimate skip), $-4$ (nonrespondent), $-5$ (out of range), $-6$ (component not applicable), $-7$ (not administered), $-8$ (unit non-response), and $-9$ (missing). The registry documents these codes and instructs the DataEngineer agent to recode all sentinel values to \texttt{NaN} before imputation. The registry also distinguishes between structural missingness (e.g., postsecondary outcomes are undefined for students who did not attend college) and random missingness, which affects whether low retention rates should trigger pipeline abort.

\paragraph{Variable semantics.} Each Tier 1 variable includes a human-written label and substantive notes. For example, \texttt{X1TXMTSCOR} is annotated not simply as ``math score'' but as ``Mathematics assessment theta (IRT) score, base year, standardized to national distribution, suitable for continuous outcome modeling.'' These annotations allow the ProblemFormulator to construct educationally meaningful research questions rather than purely statistical ones.

\paragraph{Canonical research questions.} The registry includes a curated list of canonical research questions for prediction, fairness, and policy tasks, giving the ProblemFormulator seed examples that reflect the kinds of questions the EDM community actually asks.

\subsection{Leakage Prevention}
\label{sec:leakage}

Data leakage is one of the most common and damaging errors in predictive modeling. EDM-ARS guards against it at multiple levels.

The ProblemFormulator validates that every predictor's wave precedes the outcome's wave in the registry's temporal ordering, so a research specification that attempts to use first-follow-up predictors to model base-year outcomes is rejected. The registry also flags variables that are near-duplicate composites of the outcome from the same wave (e.g., using transcript GPA to predict a closely related transcript-based outcome); the \texttt{common\_pitfalls} section of the registry explicitly warns about this pattern. At the data preparation stage, the DataEngineer is instructed to perform the train/test split before any encoding or transformation that could leak information from the test set into the training set, using a fixed random seed (\texttt{random\_state=42}) and stratification for classification outcomes. During the review stage, the Critic automatically flags any model with AUC $> 0.95$ as potentially indicative of data leakage, triggering a detailed review of the predictor set and data preparation steps. The DataEngineer also saves a \texttt{test\_protected.csv} file containing the original (pre-encoding) values of subgroup analysis variables for the test set rows, which prevents the Analyst from attempting to reconstruct categorical labels from one-hot encoded columns, a step that can introduce subtle errors.

\section{Agent Specifications}
\label{sec:agents}

Each agent inherits from a common \texttt{BaseAgent} class that provides LLM call routing (via the Anthropic SDK), sandboxed code execution (via Docker or subprocess), registry and template loading, and structured logging. All agents receive their system prompts from external YAML files and their model identifiers from \texttt{config.yaml}.

\subsection{ProblemFormulator}
\label{sec:agent_pf}

\begin{table}[H]
\centering
\small
\begin{tabular}{ll}
\toprule
Property & Value \\
\midrule
Model & Claude Sonnet (via \texttt{config.models.problem\_formulator}) \\
Temperature & 0.7 (higher for creative ideation) \\
Input & Dataset registry YAML, task template, user prompt (optional), Semantic Scholar results \\
Output & \texttt{research\_spec.json}, \texttt{literature\_context.json} \\
\bottomrule
\end{tabular}
\end{table}

The ProblemFormulator generates a structured research specification containing a research question, outcome variable, predictor set with educational rationale for each variable, target population, subgroup analysis dimensions, expected contribution, potential limitations, and a self-assessed novelty score (1--5 scale; the orchestrator rejects specifications with novelty $< 3$).

\paragraph{Literature retrieval.} The ProblemFormulator queries the Semantic Scholar API to retrieve 8--12 relevant papers for the proposed research question. Retrieved papers provide the basis for the Related Work section and novelty assessment. The agent is instructed to copy paper metadata exactly as returned by the API, never fabricating or modifying citation fields, which guards against hallucinated citations. A three-layer citation verification system inspired by AutoResearchClaw validates retrieved papers: exact title matching against the Semantic Scholar response, Jaccard similarity checking for fuzzy title matches, and optional CrossRef cross-validation for high-confidence verification.

\paragraph{Semantic Scholar fallback.} If the Semantic Scholar API returns a non-200 response or times out, the ProblemFormulator sets \texttt{literature\_context.papers} to an empty list and notes the failure. The Writer agent handles this by using placeholder citations in \texttt{[Author, Year]} format.

\paragraph{Diversity injection.} When generating multiple candidate specifications (for multi-branch exploration in future versions), the ProblemFormulator receives prior candidates as context and is instructed to produce meaningfully different research questions, not simply swapping one predictor for a synonym.

\subsection{DataEngineer}
\label{sec:agent_de}

\begin{table}[H]
\centering
\small
\begin{tabular}{ll}
\toprule
Property & Value \\
\midrule
Model & Claude Sonnet (via \texttt{config.models.data\_engineer}) \\
Temperature & 0.0 (deterministic code generation) \\
Input & \texttt{research\_spec.json}, dataset registry YAML, raw data path \\
Output & \texttt{train\_X.csv}, \texttt{train\_y.csv}, \texttt{test\_X.csv}, \texttt{test\_y.csv}, \\
& \texttt{test\_protected.csv}, \texttt{data\_report.json} \\
\bottomrule
\end{tabular}
\end{table}

The DataEngineer generates and executes Python code that performs the complete data preparation pipeline. The code follows a strict 10-step protocol. It begins by loading the raw CSV from the configured data path and selecting only the outcome and predictor columns named in the research specification. NCES sentinel codes ($-1, -4, -7, -8, -9$) are recoded to \texttt{NaN}, and rows where the outcome variable is missing are dropped (the outcome is never imputed). Per-column missingness is assessed on the remaining analytic sample, after which the train/test split is performed (80/20, stratified for classification, \texttt{random\_state=42}). Before encoding, the original categorical values of subgroup analysis variables are saved to \texttt{test\_protected.csv}. The missing data protocol (Table~\ref{tab:missing}) is then applied independently per variable, followed by one-hot encoding of categorical predictors. The final step validates all outputs against a checklist: no remaining NaN values, no zero-variance predictors, expected outcome type and range, and sample sizes reported.

\begin{table}[H]
\centering
\caption{Missing data protocol applied per predictor variable.}
\label{tab:missing}
\begin{tabular}{lp{8cm}}
\toprule
Missingness & Method \\
\midrule
$<5\%$ & Median (continuous) or mode (categorical) imputation \\
$5$--$20\%$ & Multiple imputation via \texttt{IterativeImputer} (5 iterations, \texttt{random\_state=42}) \\
$>20\%$ & \texttt{IterativeImputer} with a warning in \texttt{data\_report.json} \\
Complete-case $<60\%$ of original $n$ & \textsc{abort} (unless structural missingness) \\
\bottomrule
\end{tabular}
\end{table}

\paragraph{Code self-repair.} If the generated code fails to execute (non-zero return code), the DataEngineer enters a retry loop: the error message (stderr) is fed back to the LLM with a request to produce corrected code. Up to 3 retry attempts are permitted before the pipeline aborts.

\paragraph{Safety rules.} The DataEngineer is explicitly instructed to never use \texttt{pd.to\_numeric(errors='coerce')} on all columns (as this destroys categorical labels), to never impute the outcome variable, to always perform the split before encoding, and to never call the Anthropic API or make network requests in generated code (the sandbox has no network access).

\subsection{Analyst}
\label{sec:agent_analyst}

\begin{table}[H]
\centering
\small
\begin{tabular}{ll}
\toprule
Property & Value \\
\midrule
Model & Claude Sonnet (via \texttt{config.models.analyst}) \\
Temperature & 0.0 (deterministic code generation) \\
Input & Prepared data splits, \texttt{data\_report.json}, \texttt{research\_spec.json} \\
Output & \texttt{results.json}, tables (CSV), figures (PNG) \\
\bottomrule
\end{tabular}
\end{table}

The Analyst generates and executes a Python analysis script covering model training, evaluation, interpretability, and fairness analysis.

\paragraph{Model training.} A battery of six model families is trained with hyperparameter tuning via 5-fold cross-validation on the training set: Logistic Regression or Ridge Regression (baseline linear model), Random Forest, XGBoost (gradient boosted trees), ElasticNet (regularized linear model via SGDClassifier/SGDRegressor), Multilayer Perceptron (MLP), and a StackingEnsemble that combines all individual models as a meta-learner. SVM is intentionally excluded from the pilot due to impractical \texttt{KernelExplainer} runtime on large datasets. All stochastic operations use \texttt{random\_state=42}.

\paragraph{Evaluation.} All metrics are computed on the held-out test set. For binary outcomes: AUC, accuracy, precision, recall, F1, and 95\% confidence intervals via bootstrap. For continuous outcomes: RMSE, MAE, $R^2$, and bootstrap CIs. ROC curves are generated for all models.

\paragraph{SHAP interpretability.} SHAP values are computed using the appropriate explainer for each model family: \texttt{TreeExplainer} for tree-based models, \texttt{LinearExplainer} for linear models, and \texttt{KernelExplainer} for MLP (with a sample cap of 1{,}000 observations and \texttt{nsamples=500} to manage runtime). SHAP computation for the StackingEnsemble is skipped entirely given the complexity of explaining meta-learner predictions. A shared utility function (\texttt{analysis\_helpers.safe\_shap\_values()}) handles the inconsistency between SHAP library versions that return either arrays or lists. If SHAP computation exceeds a 600-second timeout, it is logged as a warning and skipped; the Writer notes this as a limitation.

\paragraph{Subgroup fairness analysis.} The Analyst computes per-subgroup performance metrics (AUC or RMSE) for each protected attribute specified in the research specification (typically \texttt{X1SEX} and \texttt{X1RACE}). Subgroups with $n < 50$ in the test set are flagged as unreliable. Performance gaps exceeding 5 percentage points are automatically flagged for the Critic's review. The analysis uses the pre-encoding categorical labels from \texttt{test\_protected.csv} to avoid the error-prone reconstruction of categories from one-hot encoded columns.

\paragraph{Error classification.} The Analyst includes a built-in error taxonomy that classifies execution failures (from stderr) into categories: \texttt{ImportError}, \texttt{MemoryError}, \texttt{ConvergenceWarning}, \texttt{FileNotFoundError}, \texttt{SHAPTimeout}, \texttt{ValueError}, \texttt{TypeError}, and \texttt{RuntimeError}. Each category has a targeted repair prompt that is appended to the retry message, which increases the likelihood that the LLM's corrected code addresses the specific failure mode.

\subsection{Critic}
\label{sec:agent_critic}

\begin{table}[H]
\centering
\small
\begin{tabular}{ll}
\toprule
Property & Value \\
\midrule
Model & Claude Opus (via \texttt{config.models.critic}), the most capable model \\
Temperature & 0.0 (consistent, rigorous evaluation) \\
Input & \texttt{research\_spec.json}, \texttt{data\_report.json}, \texttt{results.json} \\
Output & \texttt{review\_report.json} \\
\bottomrule
\end{tabular}
\end{table}

The Critic performs automated methodological peer review across four dimensions. The problem formulation review evaluates the research question's clarity, novelty, educational significance, and compliance with temporal ordering constraints, and checks that the novelty self-assessment is consistent with the actual contribution. The data preparation review verifies that the missing data protocol was followed, that sentinel codes were properly handled, that the train/test split is correctly sized and stratified, and that no data leakage indicators are present. The analysis review checks that all required models were trained, that evaluation metrics are computed on the test set only, that SHAP analysis was performed on the appropriate model, that confidence intervals are reported, and that suspicious metric values (AUC $> 0.95$) are flagged. The substantive review assesses whether statistical findings are connected to educational meaning rather than merely reported as numbers, evaluates whether the discussion interprets SHAP importance values in terms of educational constructs rather than variable names, and checks whether limitations are honestly disclosed.

Each dimension receives a numeric score, and the Critic produces an overall quality score and verdict. The review report includes structured issue lists with severity levels (\texttt{critical} or \texttt{major}), descriptions, recommendations, and target agents for revision. The \texttt{revision\_instructions} field maps each targeted agent to specific instructions.

The Critic uses the most capable available model (Claude Opus). This is the single most expensive LLM call in the pipeline, but it serves as the primary quality gate.

\subsection{Writer}
\label{sec:agent_writer}

\begin{table}[H]
\centering
\small
\begin{tabular}{ll}
\toprule
Property & Value \\
\midrule
Model & Claude Sonnet (via \texttt{config.models.writer}) \\
Temperature & 0.3 (controlled creativity for prose) \\
Input & All prior agent outputs, template, figures, tables \\
Output & \texttt{paper.tex}, \texttt{references.bib} \\
\bottomrule
\end{tabular}
\end{table}

The Writer compiles all pipeline outputs into a complete LaTeX manuscript. An important design decision is that the Writer does not generate LaTeX from scratch. Instead, it fills prose into a pre-defined template (\texttt{templates/paper\_template.tex}) based on the official ACM \texttt{sample-sigconf.tex}, which uses \texttt{\%\%PLACEHOLDER:SLOT\_NAME\%\%} markers for each prose section. This separation of structural LaTeX (template) from content (LLM-generated prose) greatly improves compilation reliability.

\paragraph{Paper structure.} The generated paper follows a standard structure with approximate word targets: Abstract (150--250 words), Introduction (800--1{,}200), Related Work (500--800), Methods (800--1{,}300), Results (600--1{,}000), Discussion (600--1{,}000), and Limitations (300--500).

\paragraph{Citation handling.} For each paper in \texttt{literature\_context.papers}, the Writer generates a BibTeX entry using Semantic Scholar metadata and produces \texttt{\textbackslash cite} commands keyed to Semantic Scholar paper IDs. If \texttt{literature\_context} is null (API failure), placeholder citations in \texttt{[Author, Year]} format are used.

\paragraph{UNVERIFIED flag.} If the Critic did not issue a \textsc{pass} verdict (i.e., max revision cycles were exhausted), the Writer prepends a prominently displayed warning block and appends the full Critic review report as an appendix.

\paragraph{Writing guidelines.} The Writer's system prompt includes EDM-specific writing guidelines: use ``students'' rather than ``subjects'' or ``observations''; connect every statistical finding to educational meaning; never use causal language for correlational findings (prediction $\neq$ causation); acknowledge the automated nature of the study in the methods section; and follow APA 7th edition conventions.

\section{Inter-Agent Communication}
\label{sec:communication}

Agents communicate through a shared \texttt{PipelineContext} dataclass that holds all intermediate outputs. The context is serialized to disk at each state transition for checkpoint recovery. Each agent's output follows a strict JSON schema defined in the system specification.

\paragraph{ProblemFormulator output.} A JSON object with two top-level keys: \texttt{research\_spec} (containing the research question, outcome variable, predictor set with rationales, and other specification fields) and \texttt{literature\_context} (containing retrieved papers and novelty evidence). This output is consumed by the DataEngineer (for variable selection), the Analyst (for subgroup analysis configuration), the Critic (for research quality review), and the Writer (for manuscript content).

\paragraph{DataEngineer output.} A \texttt{data\_report.json} file containing dataset statistics (original $n$, analytic $n$, train/test sizes), per-variable missingness summaries with imputation methods applied, class balance information (for classification tasks), predictor counts before and after encoding, validation status, and any warnings, along with CSV files for train/test splits and protected attribute labels.

\paragraph{Analyst output.} A \texttt{results.json} file containing per-model performance metrics with confidence intervals, the identity of the best model, SHAP importance rankings, subgroup performance breakdowns, a list of generated figures and tables, and any errors or warnings encountered during execution.

\paragraph{Critic output.} A \texttt{review\_report.json} file containing the overall verdict, quality score, dimension-specific scores and issue lists, and targeted revision instructions keyed by agent name.

All LLM responses expected to be JSON are parsed through a shared \texttt{parse\_llm\_json()} utility that strips Markdown code fences before parsing, with \texttt{JSONDecodeError} propagating to the orchestrator for pipeline abort. This utility is necessary because LLMs frequently wrap JSON output in \texttt{```json} fences despite being instructed not to.

\section{Evaluation Framework}
\label{sec:evaluation}

EDM-ARS's evaluation framework operates at multiple levels.

\subsection{Automated Quality Checks}

The pipeline enforces programmatic quality checks at every stage boundary. At the research specification stage, these checks verify that the novelty score is at least 3, that all predictors temporally precede the outcome, that no Tier 3 variables are included, that the outcome variable exists in the registry, and that the predictor set size falls between 3 and 30. At the data preparation stage, the checks verify that no NaN values remain, that no zero-variance predictors are present, that the train/test split is correctly sized (test $\geq 20\%$), that class balance is reported, and that sample sizes exceed minimum thresholds. At the analysis stage, the checks verify that all required metrics are computed, that confidence intervals are present, that SHAP analysis was completed (or timeout documented), that subgroup analysis was performed, and that no evaluation was done on training data. At the review stage, the checks verify that the Critic produced a structured review with numeric scores, a clear verdict, and actionable revision instructions where applicable.

\subsection{Critic-Based Evaluation}

The Critic agent serves as an automated peer reviewer, scoring the complete research output on a multi-dimensional rubric. The Critic's judgment is the primary quality gate: a \textsc{pass} verdict indicates that the paper meets minimum quality standards, a \textsc{revise} verdict triggers targeted improvements, and an \textsc{abort} verdict terminates the pipeline.

\subsection{Planned: Human Evaluation Protocol}

The definitive evaluation of EDM-ARS requires human expert review. The planned protocol (Phase 5 of the roadmap) involves generating $N = 10$ EDM-ARS papers on diverse HSLS:09 prediction questions, recruiting $N \geq 3$ human researchers to author papers on matched questions, conducting blinded review by EDM/LAK-familiar researchers, and comparing quality score, error rate, and time-to-completion. This evaluation has not yet been conducted.

\section{Current Limitations and Future Work}
\label{sec:limitations}

\subsection{Current Limitations}

\paragraph{Single dataset.} EDM-ARS currently supports only the HSLS:09 public-use file. The data registry, DataEngineer prompts, and validation rules are tightly coupled to this dataset's structure and conventions. Extending to additional datasets (ELS:2002, PISA 2022, ASSISTments) requires implementing a dataset adapter abstraction layer that maps each dataset's idiosyncrasies (missing codes, variable naming, temporal structure) into a common interface. This abstraction has been designed but not yet implemented.

\paragraph{Limited literature scope.} The ProblemFormulator retrieves 8--12 papers from Semantic Scholar's unauthenticated API, which provides limited coverage of the EDM literature. The system does not have access to full-text articles, cannot assess methodological quality of retrieved papers, and may miss relevant work published in venues with lower Semantic Scholar coverage. The three-layer citation verification system reduces hallucinated citations but cannot ensure comprehensive literature coverage.

\paragraph{Single research paradigm.} The current system supports only prediction tasks. Causal inference, transfer learning, psychometric modeling, and meta-analysis each require different pipeline topologies, statistical methods, validation logic, and paper structures. The roadmap describes a task-type polymorphism architecture that would support multiple paradigms through a shared orchestrator with task-specific templates.

\paragraph{Formulaic paper output.} Papers generated by EDM-ARS follow a rigid template structure with predictable narrative arcs. The Introduction always opens with why the outcome matters, the Results always report SHAP features in rank order, and the Discussion always follows the same interpretive pattern. This quality is a direct consequence of single-shot, template-based generation. Planned improvements include multi-branch exploration (generating $N$ candidate specifications and selecting the most interesting), outline-first writing (generating a narrative arc before prose expansion), and narrative archetypes (choosing from different storytelling frames based on what the data reveals).

\paragraph{No human-in-the-loop.} The current pipeline is fully automated with no interactive checkpoints. Human researchers cannot intervene after problem formulation to steer the research direction, adjust data preparation decisions, or edit the paper draft before finalization.

\paragraph{Computational cost.} A full pipeline run requires 5 or more LLM API calls (one per agent, plus retries and revision cycles), with the Critic's Opus call being the most expensive. Total cost per paper is typically \$2--\$5 USD. This is considerably lower than general-purpose systems that use tree search (AI Scientist v2) or RL optimization (CycleResearcher), but still nontrivial for researchers without institutional API budgets.

\subsection{Future Work}

We organize future work into a phased roadmap.

\paragraph{Phase 1: Refactoring for polymorphism.} Extract task type as a first-class configuration parameter. Create \texttt{TaskTemplate} and \texttt{DatasetAdapter} abstract base classes and refactor the existing prediction workflow and HSLS:09 specifics into concrete implementations. Verify that the prediction pipeline works identically after refactoring.

\paragraph{Phase 2: Findings memory and multi-branch ideation.} Implement a persistent \allowbreak  \texttt{FindingsMemory} store that accumulates across runs, enabling cumulative knowledge building. Modify the ProblemFormulator to generate $N$ candidate specifications with diversity scoring. Add a novelty dimension to the Critic's review rubric.

\paragraph{Phase 3: Causal inference task type.} Implement a \texttt{CausalInferenceTemplate} with sub-type routing (propensity score matching, inverse probability weighting, targeted maximum likelihood estimation, heterogeneous treatment effects, optimal treatment regimes). Create causal-specific agent prompts, Analyst code templates (using \texttt{econml}, \texttt{dowhy}, \texttt{zepid}), and Critic checklists (identifiability, balance diagnostics, sensitivity analysis).

\paragraph{Phase 4: Outline-first writing and narrative archetypes.} Split the Writer into an OutlineAgent (which generates a narrative arc post-analysis, selecting from archetypes such as ``The Surprising Predictor,'' ``The Fairness Audit,'' or ``The Policy Brief'') and a ProseAgent (which expands the outline into full manuscript prose).

\paragraph{Phase 5: Multi-dataset support and transfer learning.} Create dataset adapters for ELS:2002, PISA 2022, and ASSISTments. Implement a variable alignment layer for cross-dataset transfer learning. Enable cross-population and cross-wave transfer experiments.

\paragraph{Phase 6: Controlled human evaluation.} Execute the human evaluation protocol described in Section~\ref{sec:evaluation}, comparing EDM-ARS output to matched human-authored papers on quality, correctness, and efficiency metrics.

\section{Reproducibility}
\label{sec:reproducibility}

EDM-ARS is designed for straightforward deployment. This section provides the minimal instructions needed to reproduce a pipeline run.

\subsection{Prerequisites}

Running EDM-ARS requires Python 3.11 or higher, an Anthropic API key (set as the environment variable \texttt{ANTHROPIC\_API\_KEY}), and the HSLS:09 public-use CSV file placed at \texttt{data/raw/}. Docker Engine $\geq$ 24.0 is optional; the system falls back to subprocess execution if Docker is unavailable.

\subsection{Installation}

\begin{lstlisting}[language=bash]
# Clone the repository
git clone https://github.com/cgpan/edm-ars-public.git
cd edm-ars-public

# Install Python dependencies
pip install -r requirements.txt

# (Optional) Build the Docker sandbox image
docker build -t edm-ars-sandbox:latest .

# Set the API key
export ANTHROPIC_API_KEY="your-key-here"
export SEMANTIC_SCHOLAR_API_KEY="your-s2-api-key"
\end{lstlisting}

\subsection{Running the Pipeline}

\begin{lstlisting}[language=bash]
# Run with default settings (HSLS:09 prediction)
python -m src.main --dataset hsls09_public

# Run with a custom research prompt
python -m src.main --dataset hsls09_public \
  --prompt "Predict college attendance from 9th-grade indicators"

# Resume from a checkpoint after interruption
python -m src.main --dataset hsls09_public \
  --output-dir output/run_20260317_120000 --resume

# Run without Docker sandbox (subprocess fallback)
# Set sandbox.enabled: false in config.yaml
python -m src.main --dataset hsls09_public
\end{lstlisting}

\subsection{Output Structure}

Each pipeline run produces a timestamped output directory:

\begin{lstlisting}[language=bash]
output/run_YYYYMMDD_HHMMSS/
  config_snapshot.yaml    # Copy of config at run time
  checkpoint.json         # Pipeline state (for resume)
  research_spec.json      # ProblemFormulator output
  literature_context.json # Retrieved papers
  data_report.json        # DataEngineer output
  train_X.csv, train_y.csv, test_X.csv, test_y.csv
  test_protected.csv      # Pre-encoding subgroup labels
  results.json            # Analyst output
  review_report.json      # Critic output
  paper.tex               # Final manuscript
  references.bib          # Bibliography
  *.png                   # Figures (ROC curves, SHAP plots)
  pipeline.log            # Full execution log
\end{lstlisting}

\subsection{Development Commands}

\begin{lstlisting}[language=bash]
# Run tests
pytest tests/ -v

# Lint
ruff check src/ tests/

# Type check
mypy src/
\end{lstlisting}

\bibliographystyle{plainnat}

\begin{thebibliography}{10}

\bibitem[Analemma(2026)]{fars2026}
Analemma.
\newblock Introducing {FARS}: Fully Autonomous Research System.
\newblock \url{https://analemma.ai/blog/introducing-fars/}, 2026.

\bibitem[AutoResearchClaw(2026)]{autoresearchclaw2026}
Aiming Lab.
\newblock {AutoResearchClaw}: Automated research with citation verification.
\newblock \url{https://github.com/aiming-lab/AutoResearchClaw}, 2026.

\bibitem[Feng et~al.(2026)]{feng2026aletheia}
T.~Feng, T.~H. Trinh, G.~Bingham, D.~Hwang, Y.~Chervonyi, J.~Jung, J.~Lee, C.~Pagano, S.~Kim, F.~Pasqualotto, S.~Gukov, J.~N. Lee, J.~Kim, K.~Hou, G.~Ghiasi, Y.~Tay, Y.~Li, C.~Kuang, Y.~Liu, et~al.
\newblock Towards autonomous mathematics research.
\newblock \emph{arXiv preprint arXiv:2602.10177}, 2026.

\bibitem[Lu et~al.(2024)]{lu2024aiscientist}
C.~Lu, C.~Lu, R.~T. Lange, J.~Foerster, J.~Clune, and D.~Ha.
\newblock The {AI} {S}cientist: Towards fully automated open-ended scientific discovery.
\newblock \emph{arXiv preprint arXiv:2408.06292}, 2024.

\bibitem[Romero and Ventura(2020)]{romero2020edm}
C.~Romero and S.~Ventura.
\newblock Educational data mining and learning analytics: An updated survey.
\newblock \emph{WIREs Data Mining and Knowledge Discovery}, 10(3):e1355, 2020.

\bibitem[Tang et~al.(2025)]{tang2025airesearcher}
J.~Tang, L.~Xia, Z.~Li, and C.~Huang.
\newblock {AI-Researcher}: Autonomous scientific innovation.
\newblock \emph{arXiv preprint arXiv:2505.18705}, 2025.

\bibitem[Weng et~al.(2025a)]{weng2025cycleresearcher}
Y.~Weng, M.~Zhu, G.~Bao, H.~Zhang, J.~Wang, Y.~Zhang, and L.~Yang.
\newblock {CycleResearcher}: Improving automated research via automated review.
\newblock \emph{arXiv preprint arXiv:2411.00816}, 2025.

\bibitem[Weng et~al.(2025b)]{weng2025deepscientist}
Y.~Weng, M.~Zhu, Q.~Xie, Q.~Sun, Z.~Lin, S.~Liu, and Y.~Zhang.
\newblock {DeepScientist}: Advancing frontier-pushing scientific findings progressively.
\newblock \emph{arXiv preprint arXiv:2509.26603}, 2025.

\bibitem[Yamada et~al.(2025)]{yamada2025aiscientistv2}
Y.~Yamada, R.~T. Lange, C.~Lu, S.~Hu, C.~Lu, J.~Foerster, J.~Clune, and D.~Ha.
\newblock The {AI} {S}cientist-v2: Workshop-level automated scientific discovery via agentic tree search.
\newblock \emph{arXiv preprint arXiv:2504.08066}, 2025.

\end{thebibliography}

\end{document}